# Performance Evaluation of Deep Learning Models for Water Quality Index Prediction: A Comparative Study of LSTM, TCN, ANN, and MLP

Muhammad Ismail1*, Farkhanda Abbas2†, Shahid Munir Shah3†, Mahmoud Aljawarneh4†, Lachhman Das Dhomeja5†, Fazila Abbas6†, Muhammad Shoaib7†, Abdulwahed Fahad Alrefaei8†, Mohammed Fahad Albeshr9†

[1*]Department of Computer Science, Karakoram International University, Gilgit, 15100, Pakistan.
[2]School of Computer Science, China University of Geosciences, Wuhan, 430074, China.
[3]Department of Computing, Faculty of Engineering, Science, and Technology, Hamdard University, Karachi, 74600, Pakistan.
[4]Faculty of Information Technology, Applied Science Private University, Amman, 11937, Jordan.
[5]Department of Information Technology, Faculty of Engineering and Technology, University of Sindh, Jamshoro, 76080, Pakistan.
[6]Institute of Soil and Environmental Sciences, University of Agriculture Faisalabad, Faisalabad, 38000, Pakistan.
[7]State Key Laboratory of Hydraulic Engineering, Simulation and Safety, School of Civil Engineering, Tianjin University, Tianjin, 300072, China.
8Department of Zoology, College of Science , King Saud University, Riyadh, 11451, Saudi Arabia.
9Department of Zoology, College of Science, King Saud University, Riyadh, 11451, Saudi Arabia.

*Corresponding author(s). E-mail(s): muhammad.ismail@kiu.edu.pk;

Contributing authors: shamin0427@cug.edu.cn; shahid.munir@hamdard.edu.pk; ma jawarneh@asu.edu.jo;lachhman@usindh.edu.pk; fazeelaabbas802@gmail.com; xs4shoaib@tju.edu.cn; afrefaei@ksu.edu.sa; albeshr@ksu.edu.sa;
†These authors contributed equally to this work

**Abstract**

Environmental monitoring and predictive modeling of Water Quality Index (WQI) through the assessment of the water quality is an important area of research. In order to predict WQI this study utilizes four popular Deep Learning models i.e. Artificial Neural Network (ANN), Temporal Convolutional Network (TCN), Long Short-Term Memory (LSTM), and Multi-Layer Perceptron (MLP). Based on the Area Under the Operating Characteristic Curve (AUC), the per- formance of the employed models is evaluated and the results are compared. Comparative analysis showed that TCN achieved the highest AUC score i.e. 0.94 as compared to the other employed models i.e. MLP: 0.94, 'LSTM': 0.77, and 'ANN': 0.93. This comparative study provides insights into the suitability of various Deep Learning models for predicting WQI, contributing to advancements in environmental monitoring and management practices.

## 1 Introduction

Increased population, urbanization, adoption of modern life styles, and congested population structures pose problems of sewage disposal and pollution of surface waters like lakes. Natural water gets polluted because of weathering of rocks, seepage of soils,and mining processes, etc. [1]. Water quality assessment is used to assess the qualityof water based on multiple parameters such as temperature, electrical conductivity, nitrate, phosphorus, potassium, dissolved oxygen, etc. Water Quality Index (WQI) aggregates data from these parameters and produces a single numer that is helpful for the water quality assessment [2]. It facilitates a thorough judgment of water conditions in an environment and directs resource management strategies along with the appropriate treatment plan for it [3–5].

Traditionally, WQI is estimated using different mathematical procedures [6], however, recently, Machine Learning (ML) methods are used for its more feasible and cost-effective estimation [7]. Because of their robust nature to handle complex data patterns, these methods have become a viable paradigm of improved predictions. However, along with their numerous advantages, ML methods suffer from a few limitations too. These methods rely on manually extracted features from data. Furthermore, domain expertise are required to extract relevant features [8]. Also extracted features need a huge preprocessing before given to the models. This makes ML methods a highly time consuming and prone to errors process when applied for specialized applications like WQI prediction [9].

Recently emerged Deep Learning (DL) is the next level of ML that contains powerful and advanced algorithms with compelling advantages over traditional ML algorithms. DL algorithms automatically extract features form the raw input data and there is no need of manual features extraction for them. Automatically extracting features from data allow DL algorithms to provide improved predictive performance as it minimizes human biases and the possibility of the information loss produced by manual features extraction and preprocessing task.

Like the other applications areas, DL models may be a viable paradigm of WQI prediction [10]. Their robust nature to handle complex data patterns make them capable of easily capturing nonlinear characteristics hidden in water quality dynamics [11–13]. Furthermore, their capability to operate along with large-scale and high-dimensional datasets enhance the accuracy of prediction and helps in taking better decisions. By offering more accurate visions into the dynamics of water quality, applying DL techniques to WQI prediction demonstrates not only its superiority to conventional ML methods but also its potential to transform water sustainability practices and support a more resilient and sustainable environment [14, 15].

Based on the strengths mentioned above, this research employs different DL architectures for WQI prediction. Four popular DL architectures based on their capabilities and characteristics have been employed i.e. Long Short-Term Memory (LSTM), Temporal Convolutional Network (TCN), Artificial Neural Network (ANN), and Multi-Layer Perceptron (MLP). LSTM exhibits high-performance in capturing tempo-ral dependencies and sequence patterns. TCN uses convolution mechanism to extract temporal features [16–18]. ANN and MLP have the ability to intricate correlation between different water quality matrices.

The purpose of this study is to provide a comparative analysis of the aforementioned DL architectures for WQI prediction. To the best of our knowledge, none of the existing WQI prediction researches have utilized the combination of such popular DL architectures for the comparative analysis. Following are the contributions of this research:

- Employing four popular Dl architectures for WQI prediction i.e. Artificial Neural Network (ANN), Temporal Convolutional Network (TCN), Long Short-Term Memory (LSTM), and Multi-Layer Perceptron (MLP).
- Comparative analysis of the results obtained by the employed DL models

Since, each DL model can have some pros and cons. therefore, their comparative analysis will provide the trade-off between the models that can be useful to have a better WQI predictions. The remaining of the paper is organized as follows: Section 2 provides the detail on the methodology adopted in this study. This section provides detail of data acquisition, and theoretical background of DL models used in this study. Section 3 provides the detail of the results obtained and discussion on the results. Finally, section 4 presents conclusion of the study with possible future directions.

# 2 Methodology

The methodology of this study is based on data acquisition, data preprocessing, and DL models training to predict WQI for water quality assessment. Each step is detailed below.

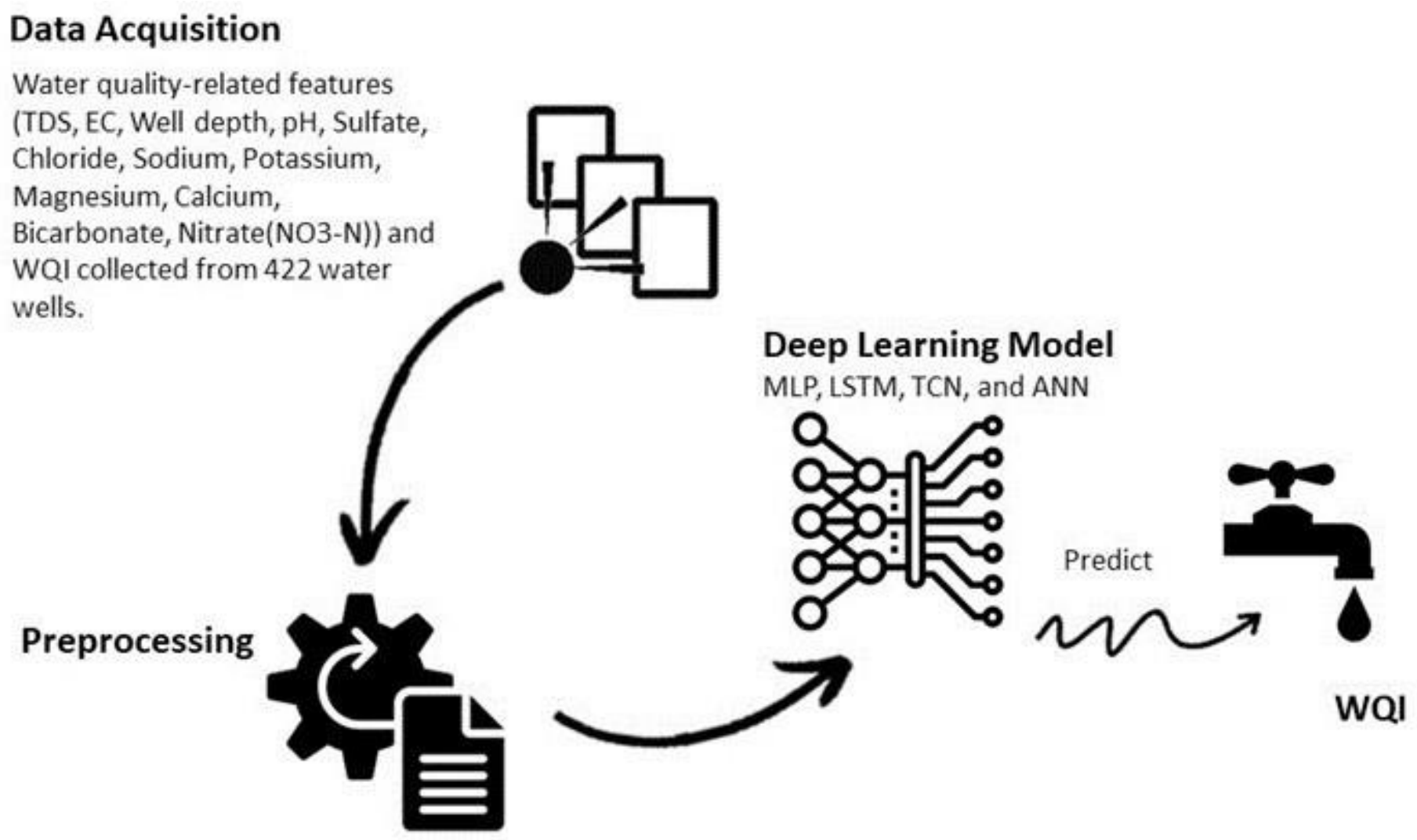

**Fig. 1** Methodology used for predicting water quality index for given input variables using deep learning techniques.

## 2.1 Data Acquisition

The dataset for this study was acquired through a proper investigation and research. Following steps were followed for this purpose i.e. study area exploration, water samples collection, features extraction, multicolinearity study of the extracted features, and finally, the dataset was collected form the explored study area. The detail of each of these steps is provided below.

### 2.1.1 Study Area Exploration

To acquire the data for our study, we selected district Mirpurkhas, situated in the Sindh province of Pakistan (refer to Figure 2). District Mirpurkhas undergoes a dry to semi-arid climate marked by extremely hot summers with temperatures frequently surpassing 40 degrees Celsius (104 degrees Fahrenheit) between April and September. The monsoons, occurring from July to September, bring substantial rainfall, offering relief from the intense heat. Winters are comparatively mild, with temperatures ranging from approximately 10 to 20 degrees Celsius (50 to 68 degrees Fahrenheit). Positionednear the Indus River in the southern part of Pakistan, Mirpurkhas is renowned for its agricultural pursuits. The region’s wells play a pivotal role in supplying groundwater for various needs, including drinking water and agricultural irrigation, supporting the local livelihoods in this semi-arid locale.

Mirpurkhas, a town in the Sindh province of Pakistan, heavily depends on well water for diverse purposes. The residents primarily use well water for drinking, agri- cultural irrigation, and domestic requirements. Wells in the area serve as the main source of groundwater, catering to the water needs of the local community. The qualityof well water in Mirpurkhas is critical for sustaining daily activities and agricultural practices. However, akin to many regions relying on groundwater, the well water qual-ity is vulnerable to contamination from sources such as agricultural runoff, industrial activities, and natural factors.

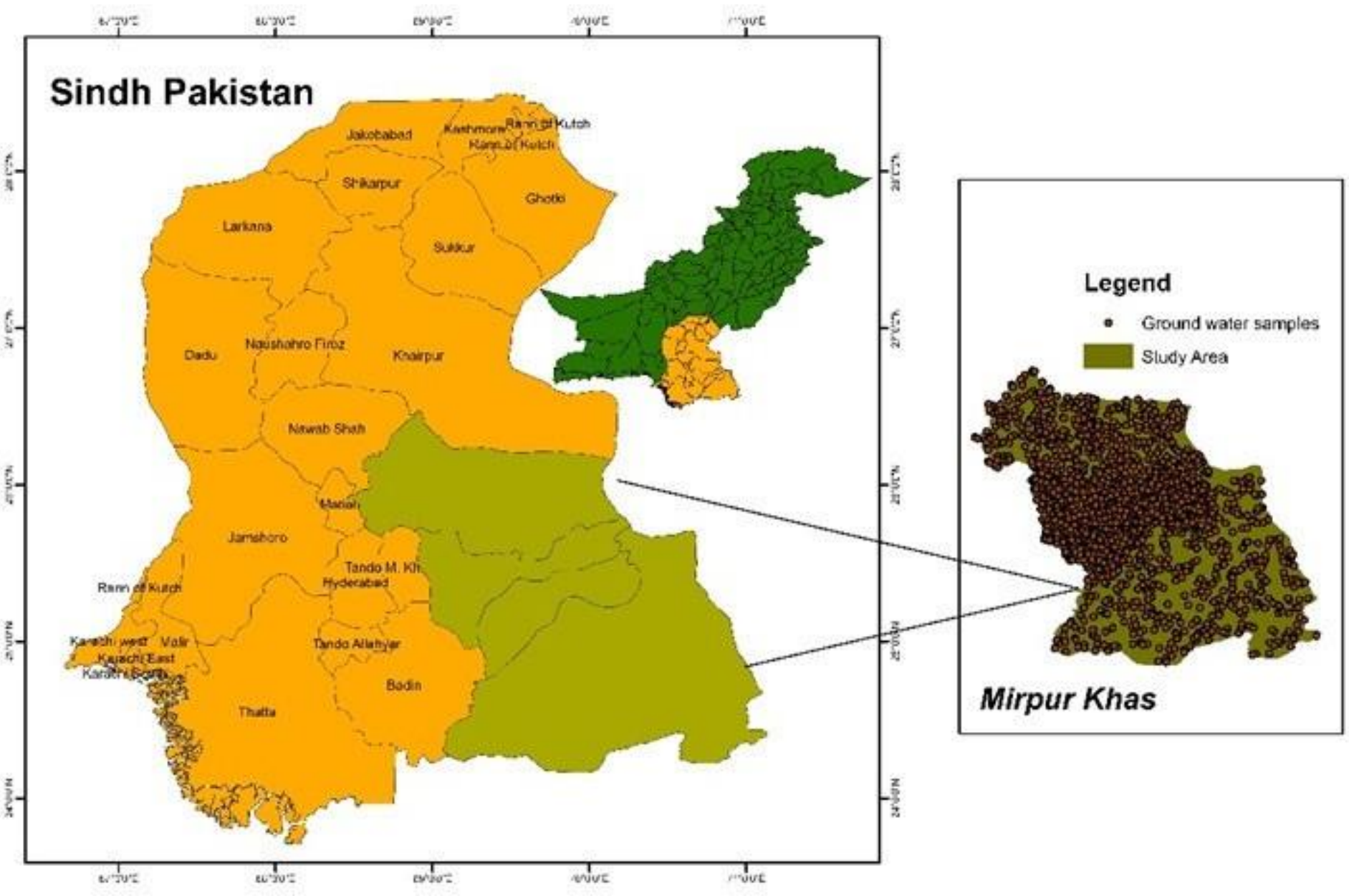

**Fig. 2** Study area and ground water sampling points.

## 2.1.2 Water Samples Collection

The dataset in form water samples was collected from different water wells of 5.7 m to 590 m depth located at various locations of the district Mirpurkhas, Sindh, Pakistan [19]. The locations of wells were extracted from places all over the city in such a manner that they represent the major place for extraction and consumption of groundwater. A total of 422 water samples were collected from the selected wells between month of April and May 2022. The collected water samples were filtered to 0.45 µm to under take further analysis.

### 2.1.3 Features Extraction and Preprocessing

Different water quality related features such as Total Dissolved Solids (TDS), Electrical Conductivity (EC), Sodium, Calcium, Magnesium, Bicarbonate, Sulfate, chloride, Potassium, Nitrate(NO3-N), pH levels, and Well depth were obtained from the collected water samples. All the features were measured and calculated using standard water testing procedures and instruments. The analysis was adhered as per the laws conveyed by the American Public Health Association (APHA). All the obtained data features were taken as independent variables while the Water Quality Index (WQI) is considered as a dependent variable. After obtaining features data, it is processed through numerous pre-processing steps such as data cleaning, handling missing values, and data normalization etc. to make the data more reliable.

## 2.2 Models Selection

After collecting and pre-processing data, four different DL models i.e. Artificial Neural Network (ANN), Temporal Convolutional Network (TCN), Long Short-Term Memory (LSTM), and Multi-Layer Perceptron (MLP) have been selected to use for WQI prediction. The purpose of using four different DL models is to provide their comparative analysis of WQI prediction, as the data of each well varies for the analysis and that each DL models have a separate purpose for the work. TensorFlow/Keras has been utilized for training the models on the collected data and for model evaluation, Area Under Operating Characteristic Curve i.e AUC has been used as a primary evaluation metrics.

### 2.2.1 The Multi-Layer Perceptron (MLP)

MLP is a fundamental type of feedforward neural network composed of multiple layers, including an input layer, one or more hidden layers, and an output layer. The input layer comprises nodes/neurons corresponding to the features or variables in the dataset. Each node represents an input feature, and these nodes pass information forward to the hidden layers. MLP consists of one or more hidden layers. Each hidden layer contains multiple neurons that perform computations on the input data. Each neuron in a hidden layer receives inputs from all neurons in the previous layer (orthe input layer for the first hidden layer). The neurons in each hidden layer apply activation functions (e.g., sigmoid, ReLU) to the weighted sum of inputs to introduce non-linearity and capture complex relationships within the data. The output layerproduces the final predictions or outputs. The number of neurons in this layer depends on the problem type: For regression tasks, there might be a single neuron providing continuous output. For classification tasks, each neuron represents a class and provides probabilities or scores for each class using activation functions like softmax (for multi-class classification) or sigmoid (for binary classification) [20, 21].

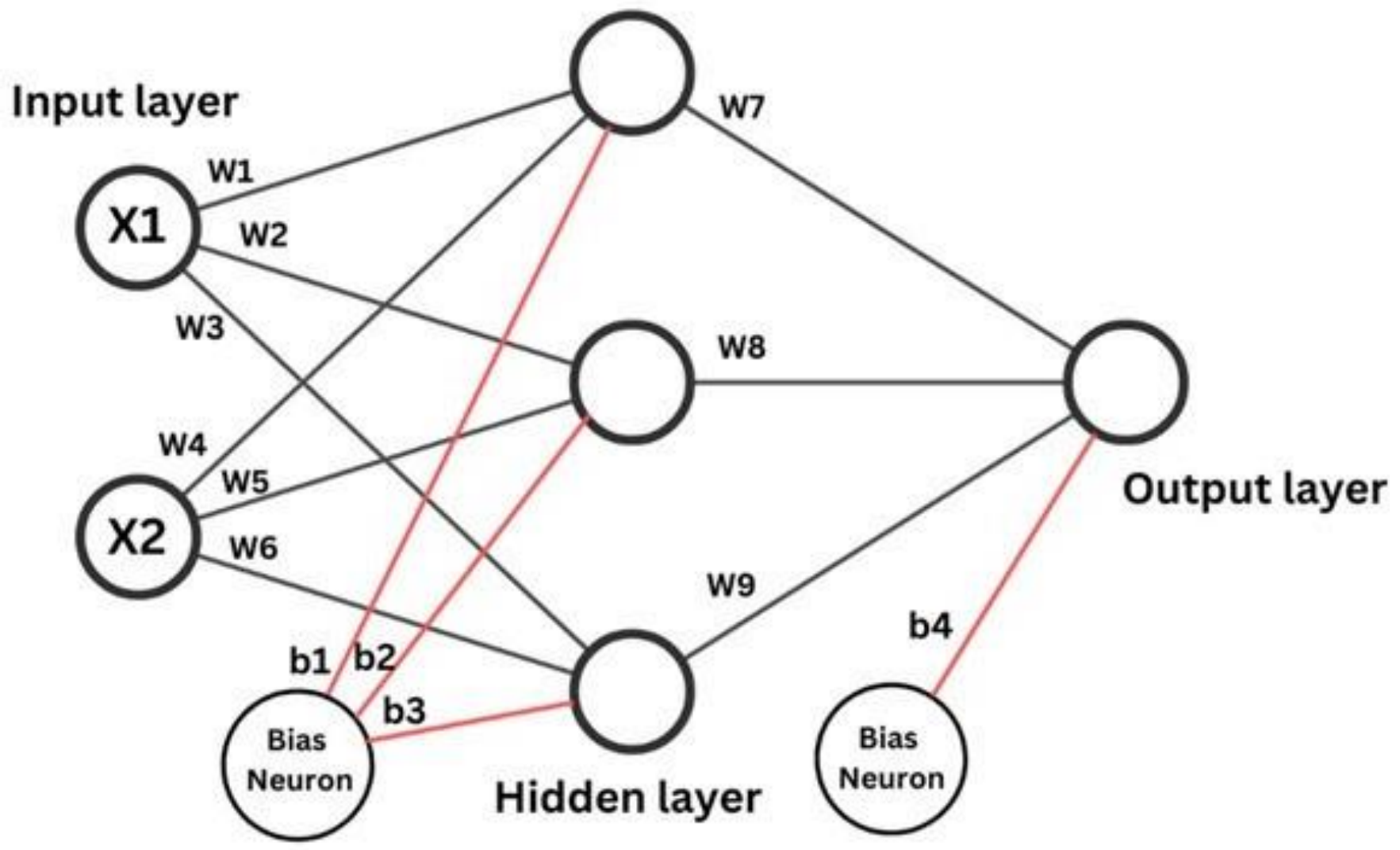

**Fig. 3** The fundamental architecture of the multi-layer perceptron ( MLP ).

Each neuron computes a weighted sum of its inputs (from the previous layer) using weights (parameters) associated with connections and adds a bias term. Mathematically, the weighted sum z for a neuron in a hidden or output layer can be represented as:

$$z = w_1x_1 + w_2x_2 + \cdots + w_nx_n + b \tag{1}$$

where w represents weights, X represents inputs, n is the number of inputs, and b is the bias term. The weighted sum is passed through an activation function (e.g., sigmoid, tanh, ReLU), transforming the result into the neuron's output For example, the output a of a neuron using the sigmoid activation function is:

$$a = 1/(1 + e^{-z}) \tag{2}$$

The output of the last layer is used for predictions or classifications. During train- ing, the network learns by minimizing a defined loss function (e.g., mean squared errorfor regression, cross-entropy for classification) using optimization algorithms like Gra- dient Descent. Backpropagation computes gradients of the loss function with respect to the network weights and biases, allowing for the adjustment of weights to minimizethe error. This basic mathematics provides insight into the structure and mathemati- cal operations within a Multi-Layer Perceptron (MLP), a foundational type of neural network used for various machine learning and deep learning tasks [22–24].

### 2.2.2 Long Short-Term Memory (LSTM)

networks are a type of recurrent neural network (RNN) specifically designed to overcome the vanishing or exploding gradient problems faced by traditional RNNs. LSTMs are capable of learning long-term dependencies and are widely used in sequence modeling tasks [25, 26]. An LSTM unit consists of a cell, three gates (input, forget, and output gates), and an internal memory state . The cell state serves as an information highway that carries information across time steps, allowing the network to retain and forget information selectively [27]. Forget Gate: Decides what information to discard from the cell state, Input Gate: Determines what new information to store in the cell state, Output Gate: Controls what information from the cell state to output as the LSTM's final prediction. The cell state runs linearly across time steps and is modified by the gates. Each gate is a sigmoid neural network layer followed by a pointwise multiplication operation and controls the flow of information by regulating how much information should be passed through. LSTMs are adept at capturing long-range dependencies in sequential data by selectively retaining and forgetting information, making them suitable for tasks involving time-series data, natural language processing, and other sequential data analysis. The architecture and gating mechanisms enable LSTMs to mitigate the vanishing or exploding gradient problem often encountered in traditional RNNs, enabling more effective learning of long-term dependencies [28–30].

The equations governing the behavior of an LSTM unit involve operations of gates and memory state manipulations. The forget gate determines what information from the previous cell state $C_{t-1}$ should be discarded. It generates a value between 0 and 1 for each element in the cell state. The equation for the forget gate $f_t$ is [31]:

$$f_t = \sigma(w_f.[h(t-1), x_t] + b_f) \tag{3}$$

Where $w_f$ are the weights for the forget gate, $h_{t-1}$ is the previous hidden state, $x_t$ is the current input, $b_f$ is the bias for the forget gate, $\sigma$ represents the sigmoid activation function. The input gate decides what new information to store in the cell state. It consists of two parts: determining the candidate values ($C_t$) and the input gate $i_t$ The equations are:

$$\tilde{C}_t = \tanh(w_c.[h_{t-1}, x_i] + b_c) \tag{4}$$

$$i_t = \sigma(w_i.[h_{t-1}, x_t] + b_i) \tag{5}$$

Where $w_c$ are the weights for the candidate values, $b_c$ is the bias for the candidate values, $w_i$ are the weights for the input gate, and $b_i$ is the bias for the input gate. The new cell state $C_t$ is a combination of the previous cell state $C_{t-1}$ that was decided to be retained (scaled by $f_t$ ) and the new candidate values scaled by the input gate $i_t$:

$$C_t = f_t.C_{t-1} + i_t.\tilde{C}_t \tag{6}$$

The output gate decides what part of the cell state should be exposed to the next hidden state $h_t$. It generates $h_t$ by passing the cell state $C_t$ through a tanh activation function and gating it with the output gate $o_t$:

$$o_t = \sigma(w_o.[h_{t-1}, x_t] + b_o) \tag{7}$$

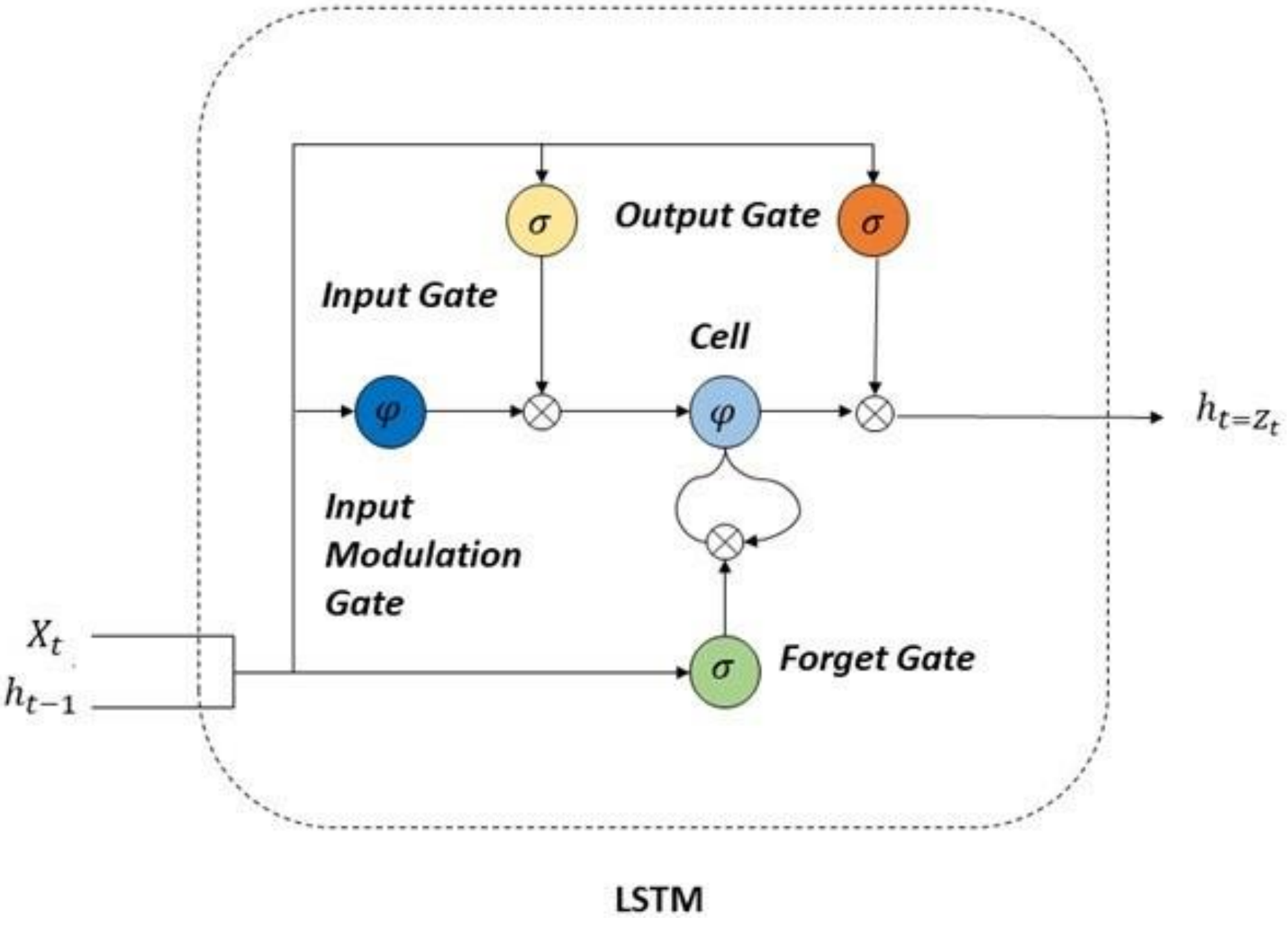

**Fig. 4** The fundamental architecture of the Long Short-Term Memory (LSTM).

$$h_t = o_t . \tanh C_t \tag{8}$$

Where $w_o$ are the weights for the output gate and $b_o$ is the bias for the output gate. These equations illustrate the operations performed by the gates in an LSTM unit to control the flow of information and update the memory state, allowing LSTMs to handle long-range dependencies in sequential data [32, 33].

### 2.2.3 Temporal Convolutional Network (TCN)

TCNs are a type of deep neural network architecture designed for sequence modeling tasks, capable of capturing long-range dependencies efficiently [34, 35]. TCNs leverage one-dimensional convolutional layers with dilated convolutions to process sequential data. The key component of TCN is the dilated causal convolution, which allows TCNs to capture long-range dependencies [36]. The mathematical representation of a dilated causal convolution operation at layer l is defined as follows [37, 38]:

$$z_l = f_l(W_l * x + b_l) \tag{9}$$

Where $W_l$ are the weights associated with the convolutional layer, $x$ is the input sequence, $*$ denotes the convolution operation, $b_l$ is the bias term and $f_l$ represents an activation function (e.g., ReLU). The dilation rate d determines the spacing between elements within the convolutional kernel, allowing for an expanded receptive field without increasing the number of parameters. The dilation factor influences the receptive field size, enabling the network to capture both short and long-term patterns. TCNs often employ residual connections to mitigate the vanishing gradient problem and enable easier training of deeper networks. The addition of residual blocks enables information flow across layers by adding skip connections to the output of convolutional layers:

$$y = x + z_l \tag{10}$$

Where $x$ is the input to the residual block and $z_l$ is the output of the convolutional layer. Temporal pooling operations (e.g., max pooling or average pooling) can be used to reduce the temporal dimensionality and capture the most salient features within the sequence [39]. The final output layer aggregates information from the preceding layers to produce predictions. For sequence classification or regression tasks, the output might pass through a fully connected layer with an appropriate activation function (e.g., softmax for classification or linear for regression) to generate the final predictions. TCNs excel in capturing temporal patterns efficiently due to the receptive field expansion achieved by dilated convolutions [**?** ]. The utilization of residual connections facilitates training deeper networks and aids in information propagation across layers, making TCNs effective for sequence modeling tasks [40, 41].

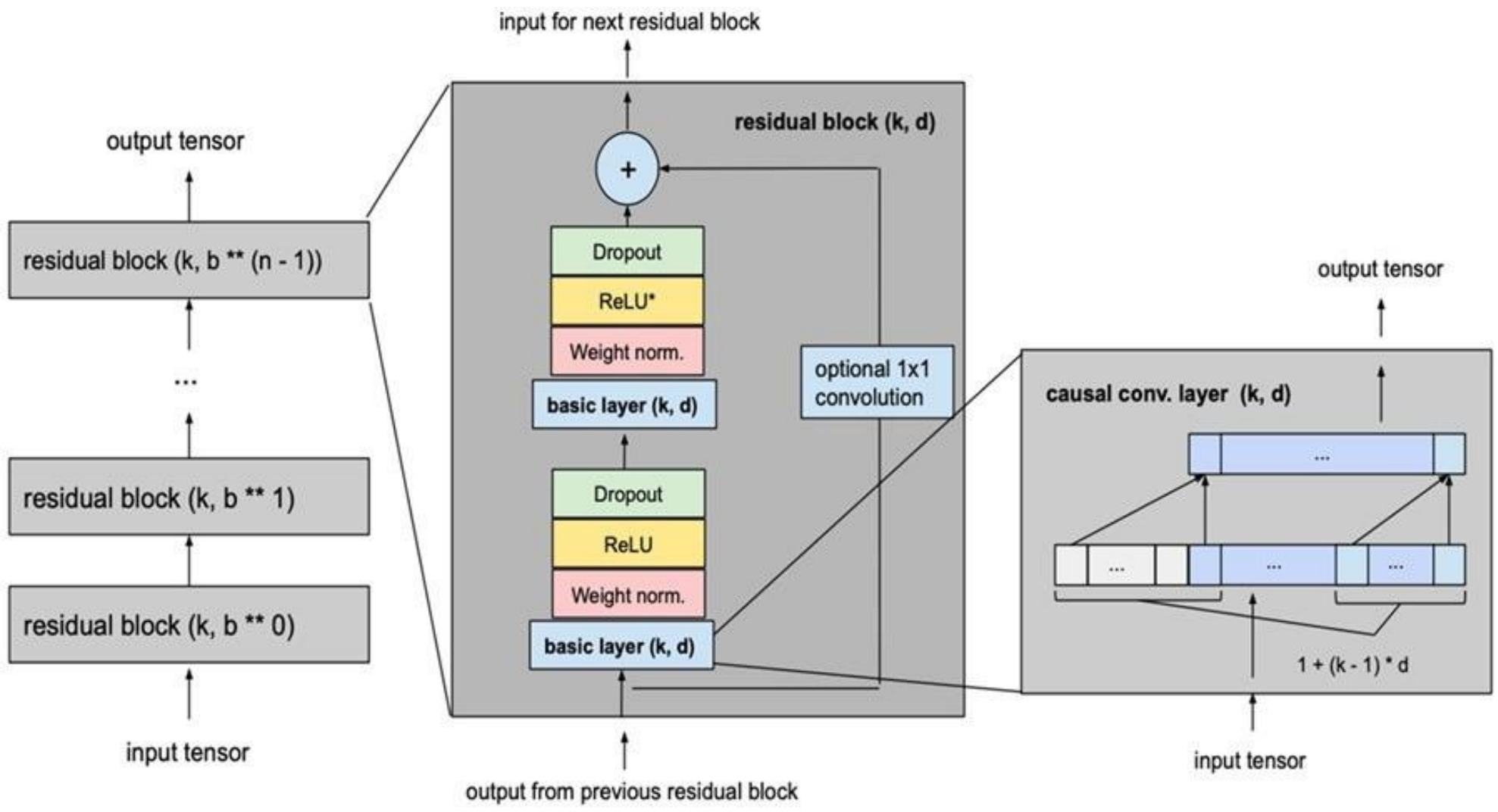

**Fig. 5** The fundamental architecture of the Temporal Convolutional Network (TCN).

## 2.3 Hyperparameter setting for models

Hyperparameter tuning is a crucial step in building effective DL models. It involves selecting the optimal values for parameters that are not learned during the training process. The hyperparameters provided in the Table 1 were obtained through a rigorous hyperparameter optimization process using a combination of grid search, random search, and Bayesian optimization techniques. These methods were applied iteratively, leveraging the computational resources available to systematically explore and evaluate the hyperparameter space. The resulting values represent the best-performing configurations for the respective DL models based on the specific dataset and performance metrics used in the optimization process.

**Table 1** Optimal Hyperparameters for Deep Learning Models Obtained Through Rigorous Hyperparameter Optimization Methods.

| Model | Hyperparameter | Best Value |
|---|---|---|
| MLP | Hidden Layer Sizes | (100, 100) |
| | Activation | relu |
| | Alpha (Regularization) | 0.001 |
| LSTM | Number of LSTM Layers | 2 |
| | Number of LSTM Units | 128 |
| | Activation | tanh |
| | Learning Rate | 0.001 |
| | Dropout Rate | 0.2 |
| TCN | Number of TCN Layers | 3 |
| | Number of Filters | 64 |
| | Kernel Size | 3 |
| | Activation | relu |
| | Learning Rate | 0.001 |
| ANN | Number of Layers | 3 |
| | Neurons per Layer | (100, 100, 50) |
| | Activation | relu |
| | Learning Rate | 0.001 |
| | Alpha (Regularization) | 0.001 |

# 3 Results and Discussion

## 3.1 Results

Table representation of the AUC (Area Under the Curve) scores obtained from training and evaluating the LSTM, TCN, ANN, and MLP models for predicting the Water Quality Index (WQI).

The AUC score measures the model's ability to distinguish between different classes, which, in the context of WQI prediction, signifies the model's capability to accurately predict various water quality levels. A higher AUC score indicates bet- ter model performance in classifying and predicting different WQI categories. Hence, ANN demonstrates the highest performance, closely followed by TCN and MLP, while LSTM shows relatively lower performance in distinguishing between different water quality levels based on the provided evaluation metric.

**Table 2** Performance Evaluation of Deep Learning Algorithms in WQI Prediction.

| Models | AUC |
|---|---|
| MLP | 0.93 |
| LSTM | 0.77 |
| TCN | 0.93 |
| ANN | 0.94 |

These AUC scores provide insights into how well each model is capturing and predicting the nuances in water quality parameters, contributing to the overall WQI. The higher AUC scores of ANN and TCN suggest their effectiveness in differentiating and predicting diverse water quality classes, which is crucial for accurate WQI assessment and environmental monitoring. Conversely, the relatively lower AUC score of LSTM indicates potential challenges in capturing and distinguishing between different water quality categories compared to the other models. The variations in the AUC scores among the MLP, LSTM, TCN, and ANN models for predicting the WQI could be attributed to several factors. The architectures of these models play a crucial role; MLPs might struggle to capture temporal dependencies present in time-series data like water quality measurements, impacting their ability to discern intricate patterns. LSTM networks excel in capturing temporal dependencies, but in this scenario, might have faced challenges in effectively leveraging such dependencies within the water quality dataset, leading to comparatively lower performance. TCNs, designed to efficiently capture temporal patterns through convolutions, could have effectively learned and utilized temporal features inherent in the water quality data, contributing to their competitive performance. ANN architectures' intricacies might have enabled them to discern complex relationships among water quality parameters, resulting in better differentiation between various water quality classes. Factors like data characteristics, model complexity, hyperparameters, and data representation methods further contribute to the varying performance, highlighting the importance of selecting models aligned with the dataset's characteristics for accurate WQI prediction.

The graph displaying predicted probabilities illustrates the likelihood of a water sample belonging to each of these classes according to the classifier's predictions. For instance The x-axis represents these five WQI classes ( Excellent, Good, Fair, Poor, Very Poor). The y-axis represents the probability of a water sample being classified into each WQI class by the classifier. Each line in the graph represents a water sample, showcasing the probabilities assigned to each WQI class for that particular sample. This visualization aids in understanding how confident the classifiers are in predicting the WQI class for each water sample. Higher probabilities for specific classessuggest higher confidence in predicting that particular water quality level. It allowsyou to assess the uncertainty or certainty of the classifier's predictions regarding the water quality classes, aiding in better comprehension of the model's confidence across different WQI levels for various water samples.

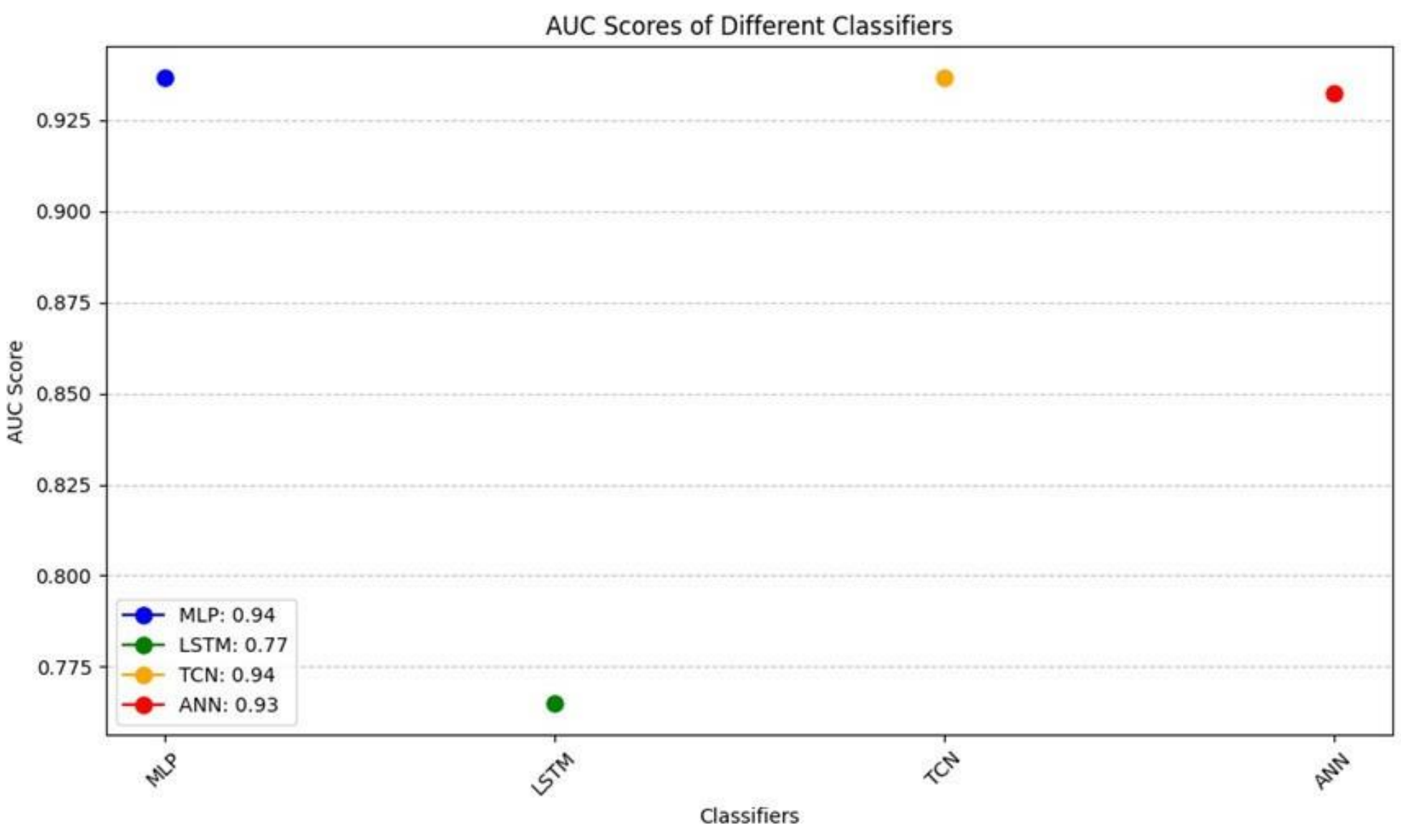

**Fig. 6** The AUC score for different deep learning classifiers used for water quality index prediction and assessment.

## 3.2 Discussion

The performance analysis of the LSTM, TCN, ANN, and MLP models for WQI pre- diction unveiled varying AUC scores, with ANN displaying the highest (0.94) followedby TCN (0.93), MLP (0.93), and LSTM (0.77). These scores depict the models' abil-ities in distinguishing between water quality categories, revealing potential strengths and limitations. Model intricacies played a pivotal role; while LSTM struggled with leveraging temporal dependencies in the water quality dataset, TCN effectively cap- tured temporal patterns through convolutions. ANN's intricate architecture enabled discernment of complex relationships among water quality parameters, contributing to its superior performance. Factors like data characteristics, model complexities, and training methodologies significantly impacted model performances. Considering their implications for WQI prediction, TCN and ANN emerged as promising models due to their adeptness in capturing temporal nuances and complex interrelationships, crucial for accurate water quality assessment. These findings have practical implications, guid-ing future environmental monitoring initiatives. Recommendations include refining model architectures and exploring ensemble methods for enhanced prediction accu-racy. Acknowledging limitations, this study underlines the significance of leveraging Deep Learning in advancing WQI prediction and environmental monitoring practices.

# 4 Conclusion and future directions

In conclusion, the comparative study evaluating LSTM, TCN, ANN, and MLP mod- els for WQI prediction highlights the significance of model architectures and data characteristics in water quality assessment. The observed AUC scores, with ANN demonstrating the highest (0.94) followed by TCN (0.93), MLP (0.93), and LSTM (0.77), emphasize the diverse capabilities of each model in capturing complex relationships and temporal patterns within water quality parameters. While LSTM faced challenges in leveraging temporal dependencies, TCN effectively captured temporal patterns, and ANN discerned intricate interrelationships, contributing to superior performance. These findings underscore the importance of aligning model intricacies with the dataset's characteristics for accurate WQI prediction. The implications for environmental monitoring are substantial, emphasizing the potential of TCN and ANN models in advancing water quality assessment practices. Recommendations include refining architectures and exploring ensemble methods to further enhance predictive accuracy. Acknowledging limitations, this study underscores the transformative role of Deep Learning in advancing WQI prediction, paving the way for improved environmental sustainability and water quality management.

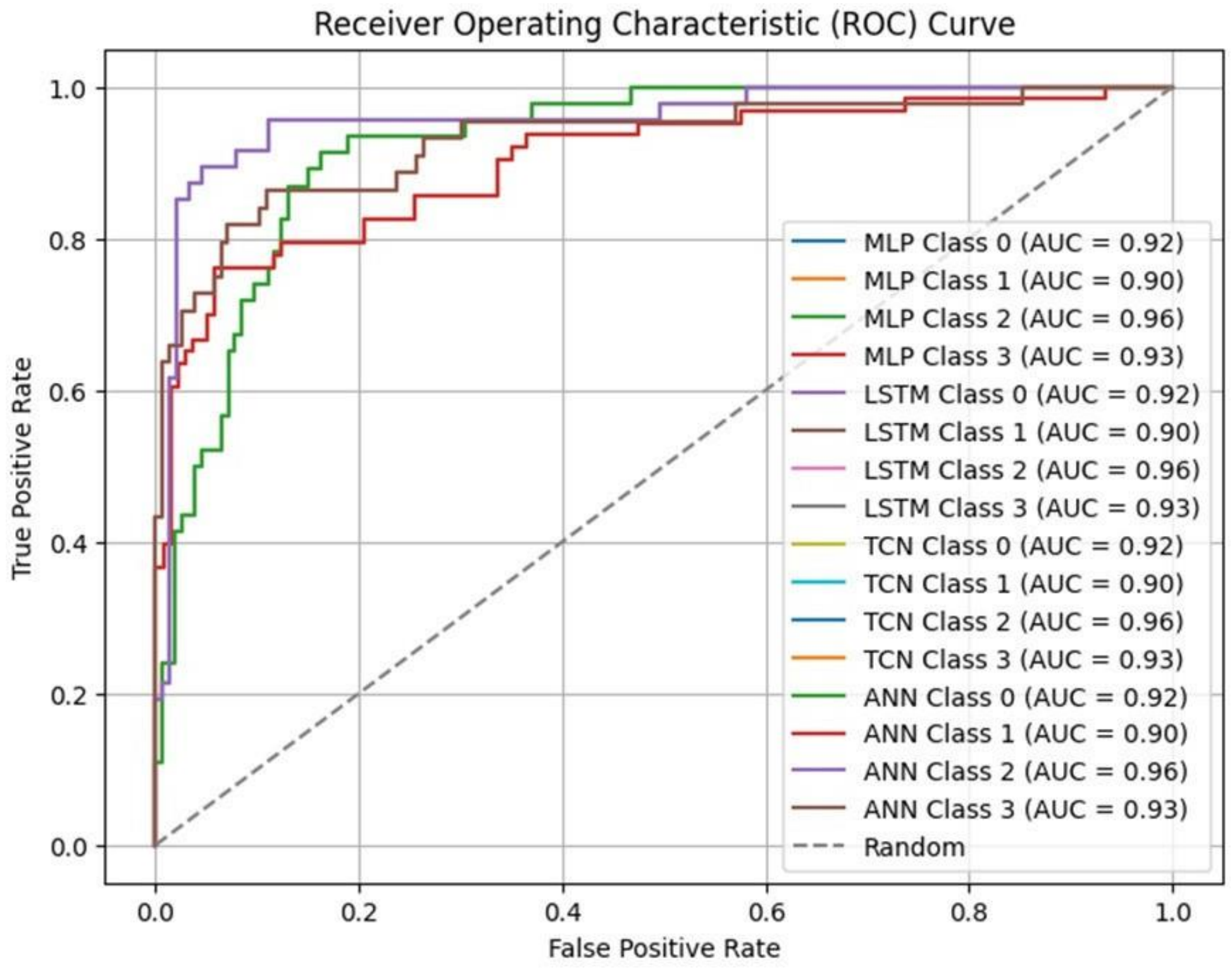

**Fig. 7** The receiver operating characteristic (ROC ) score for different deep learning classifiers used for water quality index prediction and assessment.

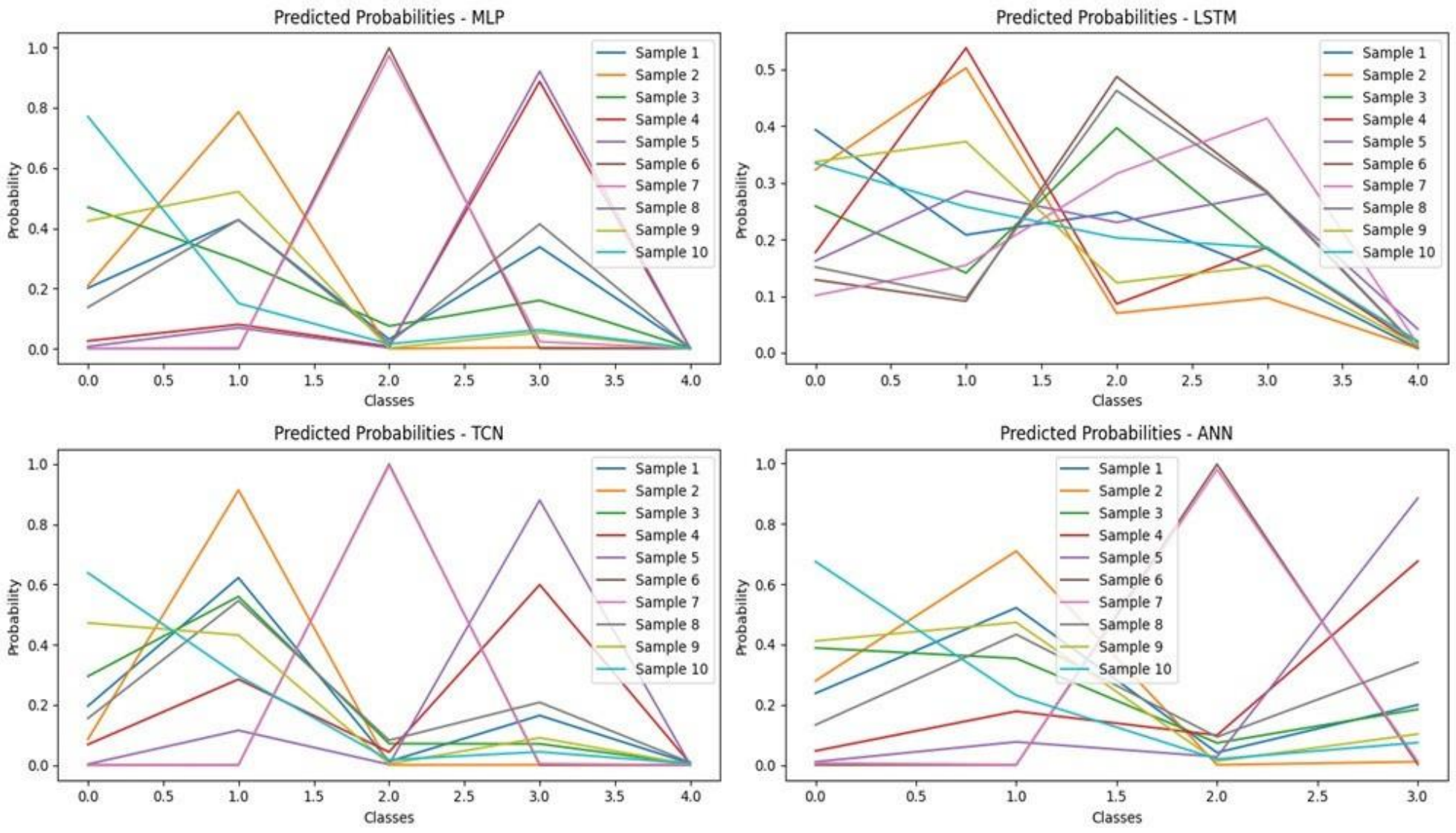

**Fig. 8** The predicted probabilities obtained from different deep learning classifiers used for water quality index prediction and assessment.